\documentclass[letterpaper, 10 pt, journal, twoside]{IEEEtran}

\IEEEoverridecommandlockouts
\usepackage{cancel}
\usepackage{graphics} % for pdf, bitmapped graphics files
\usepackage{graphicx}
\usepackage{amsmath} % assumes amsmath package installed
\usepackage{amssymb}  % assumes amsmath package installed
\usepackage{bm}
\usepackage{url}
\usepackage[table]{xcolor}
\usepackage{booktabs} % for professional tables
\usepackage{caption}% fix vertical spacing of table captions
\usepackage{multirow}
\usepackage{makecell}
\usepackage{xspace}
\usepackage{color}
\makeatletter
\DeclareRobustCommand\onedot{\futurelet\@let@token\@onedot}
\def\@onedot{\ifx\@let@token.\else.\null\fi\xspace}

\usepackage{customCommands}    % Joan Sola's macros

\def\eg{\emph{e.g}\onedot} 
\def\ie{\emph{i.e}\onedot}

\makeatother

\newcommand{\norm}[1]{\left\lVert#1\right\rVert}

\newcommand{\VPa}[1]{#1}
\newcommand{\VPr}[1]{}

\newif\ifchanges
\ifchanges
    \newcommand{\added}[1]{\textcolor{blue}{#1}} % Color for changes mode
\else
    \newcommand{\added}[1]{#1} % Default for final mode (black)
\fi

\definecolor{ao(english)}{rgb}{0.0, 0.5, 0.0}
\definecolor{blue-violet}{rgb}{0.54, 0.17, 0.89}
\definecolor{bostonuniversityred}{rgb}{0.8, 0.0, 0.0}

\title{Temporally Consistent Object 6D Pose Estimation for Robot Control
}

\author{Kateryna Zorina$^{1}$$^\ast$, Vojtech Priban$^{1}$$^\ast$, Mederic Fourmy$^{1}$, Josef Sivic$^{1}$ and Vladimir Petrik$^{1}$%
\thanks{Manuscript received: July, 16, 2024; Revised October, 5, 2024; Accepted October, 31, 2024.}%Use only for final RAL version
\thanks{This paper was recommended for publication by Sven Behnke upon evaluation of the Associate Editor and Reviewers' comments.
This work was supported by AGIMUS, euROBIN, and FRONTIER projects, funded by the European Union under GA no.~101070165, 101070596, and 101097822.} %Use only for final RAL version
\thanks{$^{1}$Kateryna Zorina, Vojtech Priban, Mederic Fourmy, Josef Sivic and Vladimir Petrik are with the Czech Institute of Informatics, Robotics and Cybernetics, Czech Technical University in Prague.
        {\tt\footnotesize name.surname@cvut.cz}}%
\thanks{$^*$Equal contribution.}
\thanks{Digital Object Identifier (DOI): see top of this page.}
}

\begin{document}

\maketitle
% TODO: use empty in submission to hide the page number
% \thispagestyle{empty}
% \pagestyle{empty}
% \thispagestyle{plain}
% \pagestyle{plain}

\begin{abstract}
Single-view RGB object pose estimators have reached a level of precision and efficiency that makes them good candidates for vision-based robot control. 
However, off-the-shelf methods lack temporal consistency and robustness that are mandatory for a stable feedback control.
In this work, we develop a factor graph approach to enforce temporal consistency of the object pose estimates. In particular, the proposed approach: (i)~incorporates object motion models, (ii)~explicitly estimates the object pose measurement uncertainty, and (iii)~integrates the above two components in an online optimization-based estimator.  
We demonstrate that with appropriate outlier rejection and smoothing using the proposed factor graph approach, we can significantly improve the results on standardized pose estimation benchmarks. 
We experimentally validate the stability of the proposed approach for a feedback-based robot control task in which the object is tracked by the camera attached to a torque controlled manipulator.
\end{abstract}

\section{INTRODUCTION}

\begin{figure}[t]
    \centering
    \includegraphics[width=\linewidth]{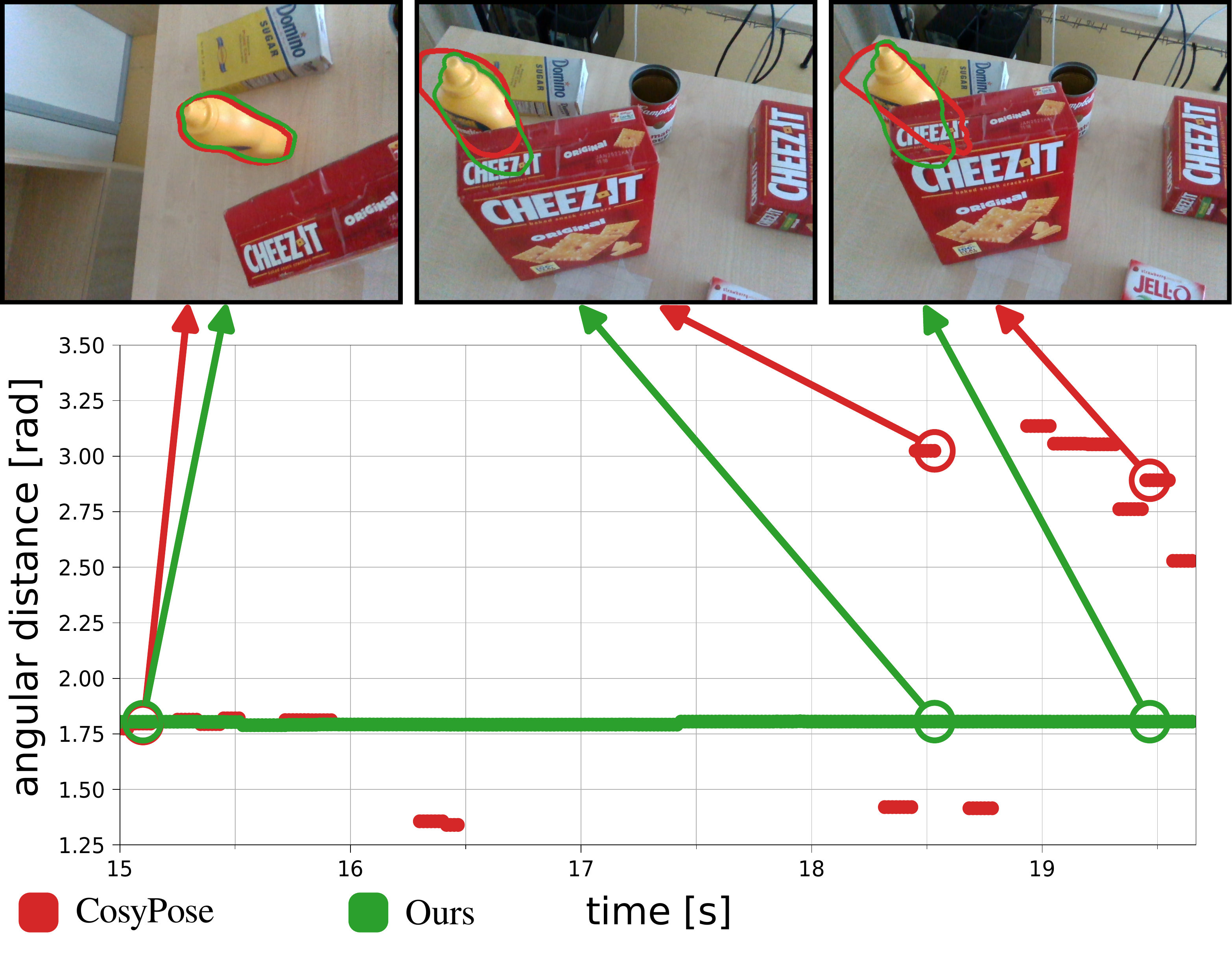}
    \caption{
    \textbf{Mustard bottle object pose estimates from images.}
    The plot (bottom) shows the angular distance between the estimated pose and a fixed reference frame. The shown objects are static, and therefore the distance should be constant.
    The red dots show the per-frame estimates computed by an object pose estimator CosyPose~\cite{labbe2020cosypose}.
    Filtered predictions computed by our method are shown in green.
    The corresponding red and green contours in the images (top) were computed by reprojecting the object model using the estimated pose.
    In the first shown frame (left), both predictions are correct and overlap.
    However, in more difficult scenarios (middle and right) the per-frame estimates (red) are incorrect  and would cause instability in the control.
    Our approach (green) is correct even in these challenging partially occluded scenarios.
    }
    \vspace{2mm}
    \label{fig:teaser}
\end{figure}

\IEEEPARstart{S}{ingle} view object pose estimation from an RGB camera has made significant progress in recent years~\cite{hodan2024bop} \eg, by using the render-and-compare approach~\cite{labbe2020cosypose,labbe2022megapose}.
Our motivation is to use object pose estimates for feedback-based robot control, for example, for visual tracking or an object hand-over from a human to a robot.
However, pose predictions are often inconsistent in time: some estimates are missing or outliers occur, as shown in Fig.~\ref{fig:teaser}.
These inconsistencies have a significant impact on the safety and robustness of feedback robot control, as incorrect pose predictions can lead to unstable behavior.
For example, incorrect pose estimates may place an object suddenly 10~cm away or incorrectly estimate that orientation suddenly changes by 180~degrees due to symmetries, and cause the controller to generate incorrectly large desired robot torques leading to dangerous motion.
Object trackers~\cite{stoiber2022srt3d,pauwels2015simtrack} provide consistent poses but may fail if objects are occluded or out of view.

To address these issues, in this paper, we build on advances in Simultaneous Localization and Mapping (SLAM)~\cite{grisetti2010tutorial} and develop a probabilistic smoothing approach to track the motion of objects based on a stream of images captured by a camera mounted on a robot arm.
The proposed approach allows us to maintain a probabilistic temporally consistent dynamic world model consisting of object poses.
Temporally consistent poses are predicted from the world model and are safe to use in the robot control loop. 
The probabilistic smoothing approach allows us to address the following challenges:
(i)~\textbf{Missing object detections} are predicted by the model (via a motion model) to maintain temporal consistency;
(ii)~\textbf{Outlier rejection} is implemented to maintain temporal consistency.
(iii)~\textbf{Multiple instances} of the same object are tracked separately in the world model so that the robot knows which instance is tracked; and
(iv)~\textbf{Discrete object symmetries} are tracked separately to predict temporally consistent poses.

In summary, this paper has the following contributions:
(i)~we present a probabilistic smoothing approach for temporally consistent object pose tracking suitable for feedback-based robot control;
(ii)~we evaluate our approach on a standard real video dataset with static objects and on synthetically rendered dataset with static and dynamic objects - we achieve superior performance on all evaluated datasets;
(iii)~we demonstrate the proposed smoothing approach in a robot object tracking application with a Franka Emika Panda manipulator - we experimentally show that our approach leads to robust tracking in situations where per-frame estimation fails.
Our code is open-source available at \url{https://github.com/priban42/temporal_pose}.

\section{RELATED WORK}

\noindent\textbf{Object pose estimation.}
% A 6D pose  is a compact spatial representation commonly used in robotics 
Model-based object pose estimation is one of the core computer-vision challenges with a wide range of applications for robotics and AR/VR~\cite{lepetit2020recent, hodavn2020bop}. The problem is most often decomposed into two stages: 2D image detection, which provides object-labeled bounding boxes and masks, followed by a pose estimation for each individual detection.
Learning methods currently dominate the standardized benchmarks for both steps~\cite{hodan2024bop}. A more recent challenge addresses generalizability to objects unseen during training, both for detection~\cite{nguyen2023cnos} and pose estimation~\cite{labbe2022megapose, nguyen2024gigaPose, ornek2023foundpose, wen2023foundationpose}. Used by some of the leading methods, the ``render-and-compare" approach~\cite{li2018deepim, labbe2020cosypose, labbe2022megapose, wen2023foundationpose} refines an initial guess by predicting object pose updates. Working with videos, this method has been shown to be competitive with the state-of-the-art single-view object pose tracking~\cite{stoiber2022srt3d, wen2020se, deng2021poserbpf, wen2023foundationpose}.

However, the single-view pose estimation problem is inherently challenging for several reasons. For RGB only methods, the geometry of pinhole projection creates a high uncertainty in the camera-to-object distance. Poses of objects can be ill-defined due to object symmetries (\eg a bottle) or partial occlusion (\eg a cup with a hidden handle). 
Higher uncertainty may also occur in real-world experiments, \eg if the model is trained with insufficient data augmentation~\cite{kendall2017uncertainties}.
With model-based object pose estimators performance improving rapidly, we propose a method that uses off-the-shelf object pose estimators for fast and robust object tracking. 

\noindent\textbf{Multiview object pose estimation.} 
In robotics, it is common to have a multi-camera setup~\cite{pauwels2015simtrack} or a camera mounted on the robot~\cite{kragic2002survey}. This setup can be leveraged by aggregating information across views and time to create a consistent estimate of both the camera/object poses and of the object shapes. Commonly used representations include parametric surfaces~\cite{nicholson2018quadricslam, yang2019cubeslam, li2021odam, laidlow2022simultaneous}, volume based representations~\cite{mccormac2018fusion++, wada2020morefusion} or latent codes~\cite{sucar2020nodeslam, landgraf2021simstack, laidlow2022simultaneous}. 

In many practical industrial scenarios, it may be reasonably assumed that object models are available before starting the tracking process. 
SLAM++~\cite{salas2013slam++} is the first depth-based object SLAM system and formulates the estimation using a probabilistic pose graph back-end. SimTrack~\cite{pauwels2015simtrack} proposes a tightly integrated RGB-D system for robot/object pose detection and tracking. 
Others directly tackle the inherent pose ambiguity of image-based pose estimation, \eg,~\cite{fu2021multi} explicitly trains a single view model that predicts a set of pose hypothesis that are resolved over different views using a max-mixture formulation. The work~\cite{merrill2022symmetry} fuses probabilistic keypoint predictions, using the known symmetries of the object. These methods require to train a dedicated "front-end" which does not clearly shows a potential for generalization. Drawing inspiration from structure-from-motion pipelines, CosyPose~\cite{labbe2020cosypose} addresses the single-view pose data association problem by designing a symmetry-aware RANSAC~\cite{fischler1981random} followed by bundle adjustment~\cite{triggs2000bundle} and is agnostic to the single view pose estimator. We propose to address the object pose ambiguities by tracking simultaneously the multiple symmetry modes of the objects in the scene.
 
\noindent\textbf{Temporally consistent moving object estimation.}
In previously mentioned SLAM-like systems, the objects are assumed to be static in the environment. A natural but challenging extension to these methods is to allow multi-object live scene reconstruction with dynamic objects~\cite{runz2017co, runz2018maskfusion, xu2019mid, li2021moltr, xu2022learning}.
To improve the geometric consistency of the scene, Dynamic SLAM~\cite{henein2020dynamic} is able to detect sparse landmarks moving with the same underlying rigid body motion model and include this information in a factor-graph-based optimization. Motion models can also provide regularization to filter-based object pose trackers, either by penalizing large pose updates~\cite{stoiber2022srt3d} or by estimating a higher-order state like the object twist~\cite{issac2016depth}. We propose to estimate the pose and twist of multiple objects using a factor graph.

\begin{figure*}[t!]
    \centering
    \includegraphics[width=\linewidth]{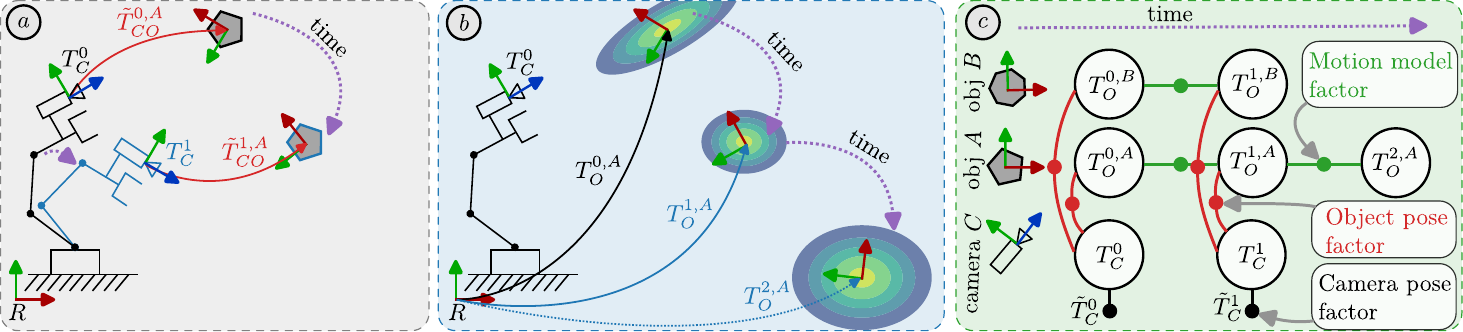}
    \caption{%
    \textbf{Overview.}
    Our goal is to estimate the poses of objects in time with respect to the reference frame $R$ as shown in figure~$a$.
    To achieve this, we use measurements at a time step~$k$ of the camera pose~$\tilde T^k_C$ and the object pose $\tilde T^{k, A}_{CO}$, where $A$ is the label of the object.
    Both objects and the robot are moving in time, as illustrated by purple arrows.
    Our approach maintains the probabilistic world representation of the object poses as visualized in figure~$b$, where the ellipsoids represent the poses uncertainty.
    This uncertainty is used to filter outliers and predict only confident poses.
    The map is maintained through the factor graph shown in figure~$c$, where the green factors represent the motion model, the red factors represent the observations of the object pose in the camera, and the black factors represent the camera pose computed by forward kinematics.
    Note that multiple objects could be tracked simultaneously, as shown by the two-object factor graph in the figure~$c$.
    Thanks to the motion model, the poses of the objects can be extrapolated to the future (figures $b$ and $c$) to resolve missing measurements due to, \eg, sudden occlusion.
    }
    \label{fig:overview}
    \vspace{-5mm}
\end{figure*}

\section{Temporally consistent pose estimation}

\noindent\textbf{Problem formulation.}
Our goal is to track the \SE(3) pose of an  object with a moving calibrated camera rigidly attached to the robot end effector, as shown in Fig.~\ref{fig:overview}-A.
To achieve that, we need to estimate the pose of $i$-th object~$T^{k,i}_{O} \in \SE(3)$ at time $k$.
The poses are expressed in the common reference frame~$R$.
Inputs to our method are the stream of images captured by the camera and the corresponding camera poses measured by the forward kinematics of the robot.
These measurements are fused into a single probabilistic estimation problem that finds the optimal trajectory of the camera and the object poses, as shown in Fig.~\ref{fig:overview}-B.
The main technical challenges are:
(i)~Achieving fast joint optimization of object poses over time while considering the uncertainties of the object pose measurements together with object motion models, which we address using a factor graph approach;
(ii)~Appropriate modelling of the measurement uncertainty of object pose, for which we develop a model that captures the difficulty of depth estimation from the RGB image;
and (iii)~Outlier rejection and data association, which we address by comparing the incoming measurements with the estimated pose distributions of the tracked objects.
The details follow.

\noindent\textbf{Factor graph.}
We formulate the temporally consistent pose estimation task as a weighted nonlinear least squares problem following the factor graph approach~\cite{dellaert2017factor}.
Under the assumption of conditionally independent measurements corrupted by Gaussian noise, the optimal sequence of object and camera poses is obtained by solving:

\newcommand{\cov}{\bfSigma}
\newcommand{\timesum}{\sum_{k=\tau-H}^\tau}
\newcommand{\timesumfromtwo}{\sum_{k=\tau-H+1}^\tau}
\begin{align}\label{equ:least_squares}
\begin{split}
    \chi^* = \argmin_\chi 
    &\underbrace{\timesum \norm{\bfr^{k}_{C}}_{\cov_C}^2}_{\text{camera pose factors}}
    +
    \underbrace{\sum_{i=1}^N\timesum \delta^{k,i}\norm{\bfr^{k,i}_{O}}_{\cov_O}^2}_{\text{object pose factors}} \\
    +
    &\underbrace{\sum_{i=1}^N\timesumfromtwo \norm{\bfr^{k-1:k,i}_{M}}_{\cov_M}^2}_{\text{motion model factors}} \, ,
\end{split}
\end{align}
where index $i$ iterates over all~$N$ objects, index $k$ represents time on the fixed time horizon~$H$ from the time of the last measurement~$\tau$,
$\bfr_X$ is the vector of residual errors weighted by covariance matrix $\cov_X$ for $X \in \{C, O, M\}$, representing the camera~$C$, the object~$O$, and motion model~$M$. 
\VPa{Term} $\delta^{k,i}$ is a binary ``occlusion" term that accounts for a missing measurement of object $i$ in frame $k$, \eg, caused by an occlusion \VPa{or a significant motion blur}.
We minimize the above cost over the set of variables~$\chi$, which consists of object and camera poses over time time, denoted as $T^{k,i}_O$ for object $i$ at time $k$ and $T^k_C$ for camera pose at time $k$.
The intuition is that
(i)~the camera pose factors regularize the camera pose to stay close to the pose measured by robot's forwards kinematics;
(ii)~the object pose factors regularize the object pose to stay close to the measured pose w.r.t. the camera, and
(iii)~the motion model factor captures the motion of the object, \ie the change of the pose and its uncertainty over time.

To account for this inequality, the residuals are scaled by covariance matrices that represent our confidence in the measurements.
The residuals are scaled by covariance matrices that represent our confidence in the measurements.
The computation of the residuals and the corresponding covariances is described next.

\noindent\textbf{The camera pose measurement factor.}
\added{We perform hand-eye calibration using the OpenCV library~\cite{opencv_library}.}
Therefore the camera pose residual can be computed by comparing the \SE(3) distance between the estimated value and the corresponding measurement, \ie, 
$
    \bfr^k_C = \Log((T^k_C)^{-1} \tilde T^k_C) \, ,
$
where symbol $\tilde{}$ represents the measurement, here computed by forward kinematics, and $\Log$ is the logarithm mapping from \SE(3) group~\cite{sola2021micro}.
The covariance of the camera pose factor is assumed to be diagonal in the form 
$\cov_C = \text{diag}(\sigma^2_{Ct}, \sigma^2_{Ct}, \sigma^2_{Ct}, \sigma^2_{Cr}, \sigma^2_{Cr}, \sigma^2_{Cr})$, where 
$\sigma^2_{Ct}$ represents the translational variance and $\sigma^2_{Cr}$ is the rotational variance.

\noindent\textbf{The object pose measurement factor.}
To estimate the pose of the object from the input RGB image we use CosyPose~\cite{labbe2020cosypose}.
CosyPose uses Mask-RCNN~\cite{he2017mask} to detect known objects bounding boxes, masks, and labels in the image.
For each image, the render-and-compare strategy is used to estimate the spatial pose of the object in the camera frame based on the 3D mesh retrieved from the database based on the predicted object identity labels.
The residual error for the $i$-th object in the $k$-th frame (time) is computed as
$
    \bfr^{k,i}_O = \Log((T^{k,i}_O)^{-1} T^{k}_C \tilde T^{k,i}_{CO}) \, ,
$
where $\tilde T^{k,i}_{CO}$ is the pose of the $i$-th object predicted by the CosyPose from the input frame $k$, $T^{k,i}_{CO}$ is the estimated temporally consistent pose of the object in frame $k$ and $T^{k}_C$ is the estimated temporally consistent pose of the camera at time $k$.

To compute the object pose residual, we need to resolve \textit{data association} between the variables (\ie object poses~$T^{k,i}_O$) and the CosyPose measurements (\ie~$\tilde T^{k,i}_{CO}$).
This is done as follows.
First, we select all variables that correspond to the predicted object label.
From this set of variables, we choose the closest one based on the Mahalanobis distance considering the estimated covariance of the measurement. 
If the translation and rotation distances are below the manually specified thresholds, denoted $\tau_\text{outlier\_t}$ and $\tau_\text{outlier\_r}$, we associate the measurement with the variable by creating a corresponding factor in the graph. 
Otherwise, a new variable is created.
This approach \added{creates a robust cost function, }enables us to filter out outliers, and track multiple instances of the same object class \added{or various discrete symmetries of the same object}.

\begin{figure}
    \centering
    \includegraphics[width=\linewidth]{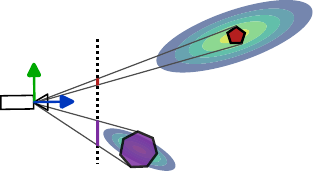}
    \caption{
    \textbf{Measurement covariance model.}
    Visualization of the translation covariance model for the object pose estimations. 
    Consider two objects (red and purple) whose projection on the image plane (dotted line) is shown in red and purple, respectively.
    The size of the covariance ellipsoid depends on the size of the object in the image plane.
    The uncertainty is higher in the direction of ray that points towards the object, mitigating the fact that the depth estimation is more difficult from monocular measurements.
    }
    \label{fig:cov_example}
    \vspace{-5mm}
\end{figure}

We observe that the covariance of the CosyPose prediction depends on the object size in the image space and that the uncertainty is higher in the direction of the ray that points from the camera towards the object center, as shown in Fig.~\ref{fig:cov_example}.
This is caused by ambiguity in depth estimation, where large changes in the depth of the object may have only a small effect on the visual appearance of the object.
Therefore, we define the object translation covariance model to deal with this increased uncertainty along the depth direction.
In particular, we define a new coordinate frame $C'$, whose position corresponds to the original camera frame and the $z$-axis points towards the object.
The translation covariance $\cov_{Ot}$ of the object pose measurement is defined in the $C'$ frame and we transform it into the object frame $O$ as:
$
    \cov_{Ot} = 
    R_{OC'}
    \cov_{C't}
    R_{OC'}^T
    \, ,
$
where $R_{OC'}$ is the rotation matrix that rotates vector from frame $C'$ to the frame $O$. 
The translation object covariance matrix in the $C'$ frame is defined as $\cov_{C't} = \text{diag}(\sigma^2_{C'xy}(n_\text{px}), \sigma^2_{C'xy}(n_\text{px}), \sigma^2_{C'z}(n_\text{px}))$, 
where the individual variances depend on the number of object pixels observed in the image~$n_\text{px}$.
We visualize the covariance model in Fig.~\ref{fig:cov_example}.
The rotational variance in the object frame is defined to be diagonal: $\cov_{Or} = \text{diag}(\sigma^2_{Or}(n_\text{px}), \sigma^2_{Or}(n_\text{px}), \sigma^2_{Or}(n_\text{px}))$.
Models for the variances $\sigma^2_{C'xy}, \sigma^2_{C'z}$, and $\sigma^2_{Or}$ are estimated on the pose estimation dataset as shown in the experiment section.
The object covariance $\cov_O$ is composed of translational and rotational covariances assuming zero correlation between them.

\noindent\textbf{Motion model factor.}
Motion model predicts the motion of the object in time.
We decoupled translation and rotation motion and compared two methods for motion prediction:
(i)~constant pose, and
(ii)~constant velocity.
The constant pose model for the residual of object~$i$  is defined as  
$
\bfr^{k-1:k,i}_{M} = \Log((T^{k-1,i}_O)^{-1} T^{k,i}_O)
$
with diaganoal covariance matrix $\cov_M = \text{diag}(\sigma^2_{Mt}, \sigma^2_{Mt}, \sigma^2_{Mt}, \sigma^2_{Mr}, \sigma^2_{Mr}, \sigma^2_{Mr}) \cdot \Delta t$, where the translation and rotation variances $\sigma^2_{Mt}$ and $\sigma^2_{Mr}$ are chosen manually, $\Delta t$ denotes the time elapsed from the previous detection of the object $i$.
With this motion model, the object pose in the world model will remain constant and its uncertainty will increase over time if no new measurements are available.

The constant velocity motion model establishes the factor on estimated derivatives of translation and rotation, from which the pose is computed via integration.
Therefore, the set of variables in Eq.~\eqref{equ:least_squares} is extended with the derivatives for each object and time stamp. The residual is computed as
$
\bfr^{k-1:k,i}_{M} = \begin{pmatrix} \bm v^{k,i} - \bm v^{k-1,i}, & \bm \omega^{k,i} - \bm \omega^{k-1,i}
\end{pmatrix}^\top,
$ where 
$\bm v^{k,i}$ and $\bm \omega^{k,i}$ represent the time derivatives of translation and rotation, respectively, for the $i$-th object at time $k$.
The covariance remains diagonal with constant variances for translation and rotation defined manually.
With this motion model, the object pose evolves based on the estimated velocity, and the uncertainty increases over time in the absence of new measurements.

\noindent\textbf{Predictions from the world model.}
Defining all the factors and solving Eq.~\eqref{equ:least_squares} gives us a probabilistic world model of all objects and camera poses in time. 
We solve the optimization for each new measurement in an iterative manner. 
\added{Incoming pose measurements are assigned to either existing or new variables in a factor graph. 
Outliers are included as potential valid measurements, but we only predict variables whose estimated uncertainty ellipsoid volume falls below manually specified thresholds $\tau_\text{pred\_t}, \tau_\text{pred\_r}$.}
If there are more identical labels that satisfy the above thresholds and whose translation distance is lower than \added{the radius of the object's bounding sphere}, we predict only the pose with the lower volume of the covariance ellipsoid, \ie the most confident track.
\added{This filtering allow us to track multiple discrete symmetries of the same object and to select only the most confident hypothesis for the robot control.}
\section{EXPERIMENTS}

\begin{figure}
    \centering
    \includegraphics[width=0.48\textwidth]{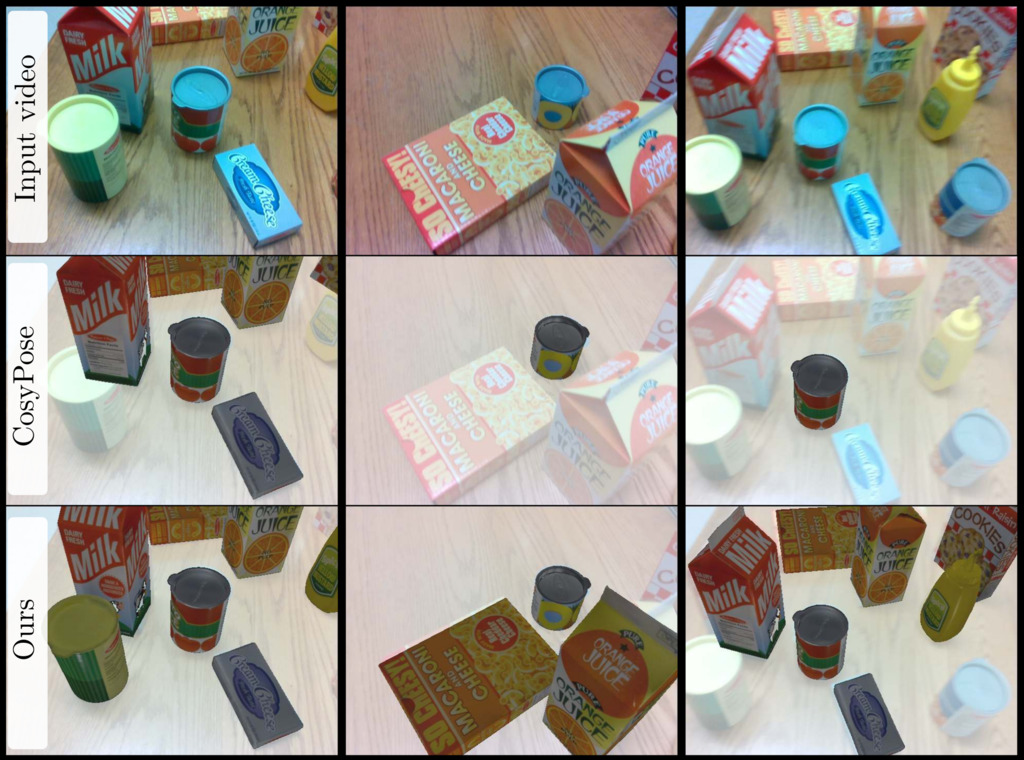}
    \caption{
    \textbf{Qualitative results on HOPE-Video.}
    Comparison between our method and CosyPose~\cite{labbe2020cosypose} on HOPE-Video sequence (the first row).
    It can be seen that some of the objects are not detected by CosyPose (the second row).
    Our temporally smoothed predictions are shown in the last row, largely mitigating the issue of missing detections.
    }
    \label{fig:hope}
    \vspace{-5mm}
\end{figure}

\begin{figure}
    \centering
    \includegraphics[width=0.48\textwidth]{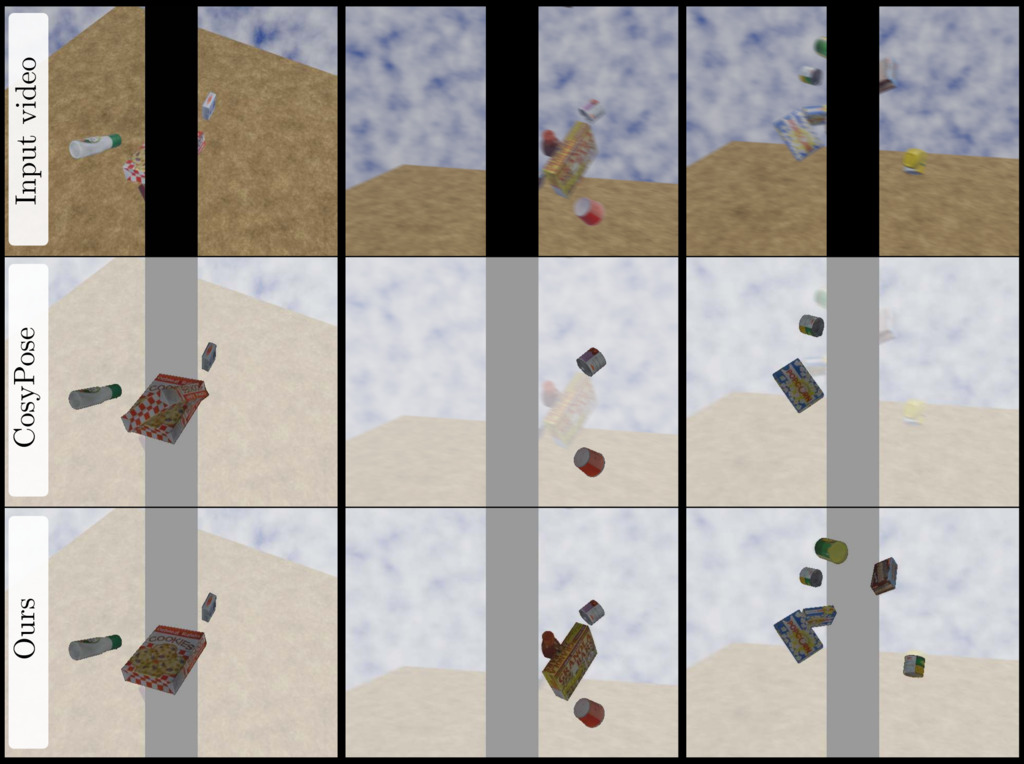}
    \caption{
    \textbf{Qualitative results on SynthHOPEDynamic.}
Comparison between our method and CosyPose~\cite{labbe2020cosypose} on \textit{SynthHOPEDynamic} sequence shown in the first row.
The center of the frame is occluded by a black rectangle, and some of the frames are artificially blurred in the input video.
It can be seen that some of the objects are not detected by per-frame CosyPose shown in the second row (\eg, frames 2 and 3) or that some outliers are detected (\eg, rotated cookie box in frame 5).
Our temporally smoothed predictions (the last row) largely mitigate missing detections and outliers.
    }
    \label{fig:synth}
    \vspace{-5mm}
\end{figure}

\noindent\textbf{Datasets.}
\added{Four} datasets were used for the quantitative evaluation: (i)~Household Objects for Pose Estimation (HOPE-Video)~\cite{tyree2022hope} dataset\added{, (ii) YCB-V~\cite{calli2015ycb, xiang2017posecnn}  and (iii})~two synthetically rendered datasets created via Blender~\cite{blender}.
Only RGB images are considered for all datasets.
HOPE-Video contains 10 video sequences captured by a moving camera observing a static scene with 5-20 objects placed on a desk.
The video is recorded by a robot equipped with a RealSense camera; an example of the video sequence is shown in Fig.~\ref{fig:hope}.
\added{YCB-V test set contains 12 video sequences captured by a moving camera observing a static scene with a subset of 21 of the YCB~\cite{calli2015ycb} objects.}

\added{The remaining datasets are synthetically rendered using HOPE objects~\cite{tyree2022hope}.}
To address the real-to-sim comparison, we first rendered 10 video sequences with static objects placed on the desk in a setup similar to the HOPE-Video datset.
We refer to this dataset as \textit{SynthHOPEStatic}.
Dynamical dataset \textit{SynthHOPEDynamic} is composed of 5-10 objects moving on randomly sampled trajectories.
The trajectories are obtained by randomly sampling poses in \SE(3) that are connected by the Cartesian dynamical movement primitives~\cite{ude2014orientation} with randomly sampled weights and initial and goal velocities.
The camera is also moving on a random trajectory and motion blur is applied to random frames.
To simulate challenging occlusions, a uniform color box is rendered in front of the camera.
In total, 10 video sequences are rendered for the \textit{SynthHOPEDynamic} dataset.
An example of the synthetic dataset is shown in Fig.~\ref{fig:synth}.
\added{In total, we have three static datasets depicting stationary objects and one dynamic dataset with moving objects.}

\noindent\textbf{Metrics.} 
To measure performance, we calculated the average recall and average precision for the three datasets.
For average recall, we rely on error metrics, which are commonly used in the BOP object pose estimation challenge~\cite{hodan2018bop}.
Recall is averaged across several thresholds and across three different metrics: 
(i)~Visible Surface Discrepancy (VSD),
(ii)~ Maximum Symmetry-Aware Surface Distance (MSSD),
and (iii)~Maximum Symmetry-Aware Projection Distance (MSPD).
See~\cite{hodavn2020bop} for details on these metrics and thresholds.
For precision, we used the same metric (\ie, VSD, MSSD, and MSPD) and the same thresholds as used for the recall computation.
Recall penalizes missing object detections and object pose estimates, while precision penalizes incorrect object detections and object pose estimates.
Only objects that are at least partially visible in the image are considered in the evaluation; \ie, the number of visible pixels is at least~5\%~of the size of the full object projection.

\begin{table}[b!]
    \centering
    \vspace{-5mm}
    \caption{
    \added{\textbf{Comparison of different covariance models.}
    Average Recall and Average Precision are computed by considering all frames of the video and all objects that are visible in the image with at least 5\% of the object size.
    The highest values are shown in bold.}
    }
     \label{tab:cov_ablation}
    \begin{tabular}{ccccc} \toprule
    
        \added{Decoupled} & \added{Visibility dependent} & \added{frame $C'$} & \added{recall} & \added{precision} \\ \midrule
        \added{\checkmark} & \added{\checkmark} & \added{\checkmark} & \added{\textbf{0.571}} & \added{\textbf{0.609}}\\
        \added{\checkmark} & \added{\texttimes} & \added{\checkmark} & \added{0.570} & \added{0.608} \\
        \added{\checkmark} & \added{\checkmark} & \added{\texttimes} & \added{0.531} & \added{0.574} \\
        \added{\texttimes} & \added{\checkmark} & \added{N/A}  & \added{0.483} & \added{0.549} \\
        \added{\texttimes} & \added{\texttimes} & \added{N/A}  & \added{0.498} & \added{0.542}\\        
    \bottomrule
\end{tabular}
\end{table}

\noindent\textbf{Measurement covariance estimation.}
We empirically observe that translation measurement uncertainty is bigger in the direction of ray pointing towards the object of interest (\ie, standard deviation~$\sigma_{C'z}$) and that it depends on the size of the object in the image space, as shown in Fig.~\ref{fig:cov_example}.
We propose to model the dependence of the standard deviation on the number of visible pixels of the object as an exponential function of the form: $\sigma(n_\text{px}) = a \exp(-b n_\text{px})$,
where $a$ and $b$ are parameters fitted separately for the translation $xy$, the translation $z$ (\ie, depth) and the rotation. 
Translation uncertainties are estimated in the coordinate frame whose $z$-axis points toward the center of the object, while rotation uncertainties are estimated in the object coordinate frame.
We used the Hope-Video dataset to estimate these uncertainties. 

\begin{figure}[t!]
    \centering
    \includegraphics[width=\linewidth]{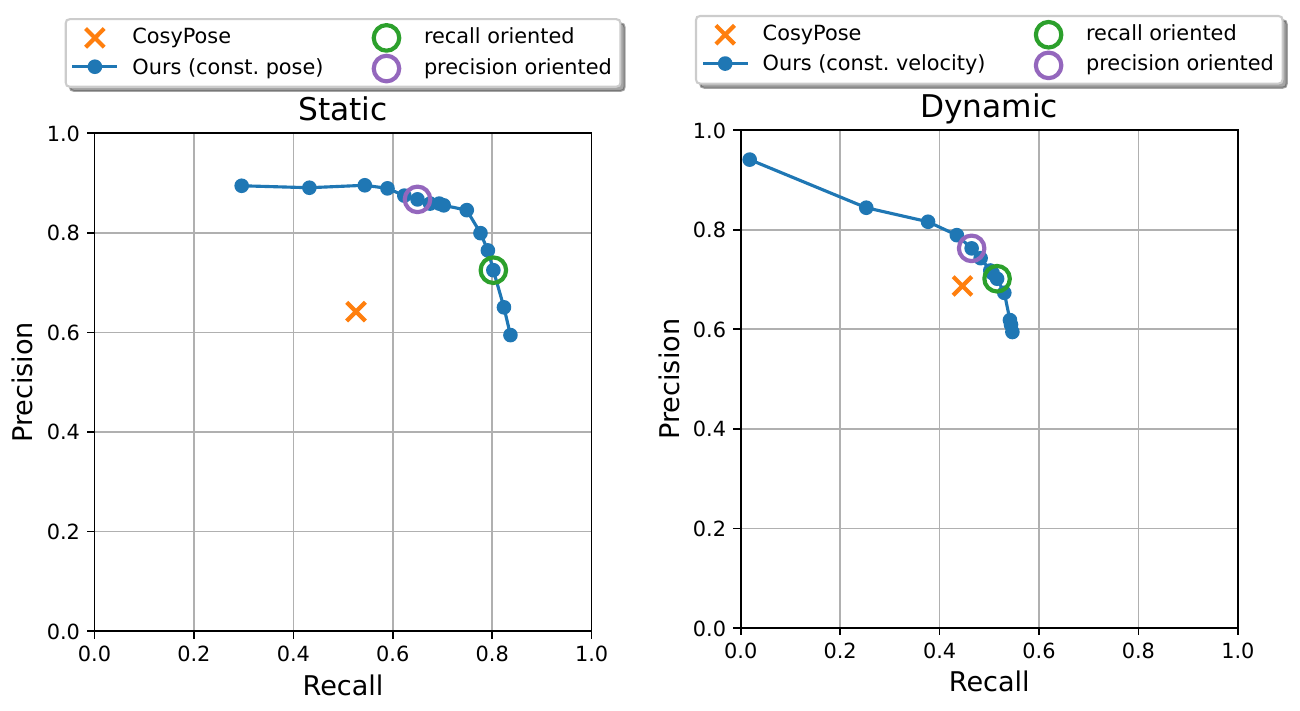}
    \caption{
    \textbf{Ablation study} for the constant pose motion model (left) evaluated on three scenes of the static synthetic dataset and for the constant velocity motion model (right) evaluated on three scenes of the dynamic synthetic dataset.
    The precision-recall trade-off is controlled by hyperparameters of our model.
    Recall oriented parameters are selected such that recall is maximal and precision is at least at CosyPose level. The precision oriented parameters are chosen analogously.
    }
    \label{fig:ablation}
    \vspace{-5mm}
\end{figure}

\begin{table*}[t!]
    \centering
    \caption{BOP Average Recall and Average Precision evaluated on three video datasets by considering all frames of the video and all objects that are visible in the image at least 5\% of the object size. "Recall-oriented" and "precision-oriented" refer to different configurations aimed at maximizing average recall or precision while ensuring that the average of the other metric is at least as good as CosyPose.
    Terms "const. pose" and "const. velocity" denote different motion models.
    The "Short-horizon" baseline refers to our method modified to use only the last 3 frames.
    We present recall and precision averaged across VSD, MSSD and MSPD metrics.
    The best results for recall and precision are shown in bold.} \label{tab:bop_evaluation}
    \begin{tabular}{lcccccccc} \toprule
    
        & \multicolumn{2}{c}{HOPE-Video} &  \multicolumn{2}{c}{\added{YCB-V}} & \multicolumn{2}{c}{SynthHOPEStatic} & \multicolumn{2}{c}{SynthHOPEDynamic} \\
        Method & recall & precision & \added{recall} & \added{precision} & recall & precision & recall & precision \\ \midrule
        CosyPose~\cite{labbe2020cosypose} & 0.39 & 0.57 & \added{0.81} & \added{0.72} & 0.53 & 0.69 & 0.44 & 0.66 \\
        Short-horizon & 0.40 & 0.59 & \added{0.81} & \added{0.72} & 0.54 & 0.75 & 0.40 & 0.71 \\
        Ours (const. pose, recall-oriented) & \textbf{0.57} & 0.61 & \added{\textbf{0.82}} & \added{0.77} & \textbf{0.74} & 0.85 & \multicolumn{2}{c}{\textit{not applicable}} \\
        Ours (const. pose, precision-oriented) & 0.43 & \textbf{0.65} & \added{0.81} & \added{\textbf{0.80}} & 0.63 & \textbf{0.89} &  \multicolumn{2}{c}{\textit{not applicable}} \\
        Ours (const. vel., recall-oriented) & 0.47 & 0.60 & \added{0.80} & \added{0.79} & 0.58 & 0.79 & \textbf{0.45} & 0.76          \\
        Ours (const. vel., precision-oriented) & 0.43 & 0.64 & \added{0.81} & \added{0.77} & 0.53 & 0.83 & 0.41 & \textbf{0.79} \\       
    \bottomrule
\end{tabular}
\vspace{-5mm}
    
\end{table*}

% HopeVideo, const pose, recall-oriented params:  0.57/0.61
% HopeVideo, const pose, precision-oriented params: 0.43/0.65
% HopeVideo, const vel, recall-oriented params: 0.47/0.60
% HopeVideo, const vel, precision-oriented params: 0.43/0.64

% YCBV, pose, recall-oriented params: 0.82/0.77
% YCBV, pose, precision-oriented params: 0.81/0.80
% YCBV, vel, precision-oriented params: 0.80/0.79
% YCBV, vel, recall-oriented params: 0.81/0.77

% SynthStatic, pose, recall-oriented params: 0.74/0.85
% SynthStatic, pose, precision-oriented params: 0.63/0.89
% SynthStatic, vel, recall-oriented params: 0.58/0.79
% SynthStatic, vel, precision-oriented params: 0.53/0.83

% SynthDynamicOcclusion, vel, recall-oriented params: 0.45/0.76
% SynthDynamicOcclusion, vel, precision-oriented params: 0.41/0.79

\noindent\textbf{Ablation study.}
Several thresholds need to be tuned for the proposed filtering method.
We manually set \added{horizon $H$ to 30~frames corresponding to 1s in our setup,} the outlier prediction thresholds $\tau_\text{outlier\_t}$ and $\tau_\text{outlier\_r}$ are set to 100~mm and 10$^\circ$.
The prediction thresholds $\tau_\text{pred\_t}$, $\tau_\text{pred\_r}$, \added{and the variances of the motion models} were chosen based on the ablation study in which we evaluated the precision-recall curve for various values of these hyperparameters.
Subsets containing three scenes from our synthetic datasets were used to select thresholds for the constant pose model (subset of \textit{SynthHOPEStatic}) and for the constant velocity model (subset of \textit{SynthHOPEDynamic}).
The result of the ablation is shown in Fig.~\ref{fig:ablation}; we use it to select two sets of thresholds for each motion model.
These sets of thresholds correspond to recall- and precision-oriented parameters as shown in Fig.~\ref{fig:ablation}.

\noindent\added{\textbf{Covariance models.} In addition to threshold selection, we conducted an ablation of different variants of covariance models to evaluate their effect on performance.
The models were evaluated on the \textit{HOPEVideo} dataset using the constant pose motion model and recall-oriented parameters.
The results are shown in Tab.~\ref{tab:cov_ablation}.
We ablate various aspects of the model: isotropic vs. decoupled along the $x$, $y$, and $z$ axes,  constant vs. visibility dependent, and expressed in camera frame vs. in rotated frame~$C'$.
The results show that our proposed covariance model has the best performance, indicating that the rotated camera frame $C'$ is important for accurate modeling of the covariance.
In contrast, the dependency on the object's visibility in the image shows only a minor improvement. 
}

\begin{table}[b]
    \centering
    \vspace{-5mm}
    \caption{
    \added{\textbf{Comparison to state-of-the-art methods.} 
    }
    }
     \label{tab:slam_comparison}
    \begin{tabular}{lcc} \toprule
    
        \added{Method} & \added{ADD-S} & \added{ADD(-S)} \\ \midrule
        \added{CosyPose~\cite{labbe2020cosypose} }& \added{0.9} & \added{0.84} \\ 
        \added{Merrill et al.~\cite{merrill2022symmetry}} & \added{0.9} & \added{0.85} \\
        \added{Xu et al.~\cite{xu2022rnnpose}} & - & \added{0.83} \\
        \added{Di et el.~\cite{di2021so}} & \added{0.91} & \added{0.84} \\ 
        \added{\textbf{Ours}} & \added{\textbf{0.94}} & \added{\textbf{0.9}} \\
    \bottomrule
\end{tabular}    
\end{table}
\noindent\textbf{Quantitative evaluation.}
We evaluated the performance of our method on the \added{four} datasets mentioned above.
The results are summarized in Tab.~\ref{tab:bop_evaluation}.
Two baselines are considered: (i) per-frame CosyPose~\cite{labbe2020cosypose} and (ii) short-horizon filtering, in which only the last three frames were used for our method.
For static object datasets (\ie \textit{HOPE-Video}\added{, \textit{YCB-V}} and \textit{SynthHOPEStatic}) both constant pose and constant velocity motion models are evaluated.
It can be seen that our methods outperformed the baselines in recall (the recall-oriented variant) while achieving comparable precision.
Similarly, for the precision-oriented variant, we outperform the baselines in precision while achieving a comparable recall.
The precision-recall trade-off can be controlled by the hyperparameters.
The constant pose motion model achieved better performance than the constant velocity motion model as it has a stronger prior about the motion of the objects.
For the dynamic object dataset, we evaluated the constant velocity motion model.
We outperform the baselines in a similar manner.

Our approach outperforms both the per-frame CosyPose and the short-horizon smoothing baselines.
Our results show that longer horizon and smoothing can be used to control recall-precision trade-off and therefore it can be tuned for robustness that is required for feedback robot control.

\begin{table}[b]
    \centering
    \vspace{-5mm}
    \caption{
    \added{\textbf{Sensitivity to backbone.} We compare our approach using different backbones on the \textit{HOPEVideo} dataset.
    }}
     \label{tab:backbone_comparison}
    \begin{tabular}{lcc} \toprule
    
        \added{Method} & \added{recall} & \added{precision} \\ \midrule
        \added{CosyPose} & \added{0.39} & \added{0.57} \\
        \added{Ours (CosyPose backbone)} & \added{0.57} & \added{0.61} \\ \midrule
        \added{MegaPose} & \added{0.37} & \added{0.54} \\
        \added{Ours (MegaPose backbone)} & \added{0.54} & \added{0.61} \\     
    \bottomrule
\end{tabular}
\end{table}

\noindent\added{\textbf{Comparison with state-of-the-art.}
To further validate our approach, we extend the evaluation to the \textit{YCB-V} dataset, utilizing the ADD-S and ADD(-S) metrics, which are commonly used to assess object pose estimation accuracy. In Tab.~\ref{tab:slam_comparison},  we present a comparison of our method against several state-of-the-art approaches on this dataset. Our method consistently outperforms the reported techniques.
}

\noindent\added{\textbf{Sensitivity to pose estimation backbone.} 
In Tab.~\ref{tab:backbone_comparison}, we compare our method's performance using different pose estimation backbones.
Specifically, we evaluate CosyPose and MegaPose as the underlying pose estimation backbones and then compare them with our approach.
The results show that our method improves both recall and precision metrics, regardless of the chosen backbone. This demonstrates that our approach is not tightly coupled to a specific pose estimation model and can be effectively integrated with various state-of-the-art methods without significant performance degradation.}

\noindent\textbf{Qualitative robotic experiment.}
To validate the stability of the proposed filtering method, we performed several robotics experiments.
For all experiments, we used a Franka Emika Panda robot equipped with a calibrated RealSense D435 camera attached to its end-effector. 
The camera produces a 60~Hz RGB video stream with a resolution of 640x480 pixels.  
We conducted the following robotic experiments to demonstrate the advantages of the method:
(i)~\textbf{Static scene} objects pose estimation in which the robot is guided by a human hand and, while moving, it estimates the poses of objects that are statically placed in front of the robot;
(ii)~\textbf{Dynamic scene} object pose estimation, where the robot remains static and estimates the poses of objects that are moved by a human; and
(iii)~\textbf{Dynamic object tracking} where the robot maintains constant pose with respect to a target tracked object.
In the first two experiments, the robot is not controlled on the basis of the predicted poses and our method can be applied directly.
Please, see the supplementary video for the recording of the experiments.

\begin{figure}[t]
    \centering
    \includegraphics[width=\linewidth]{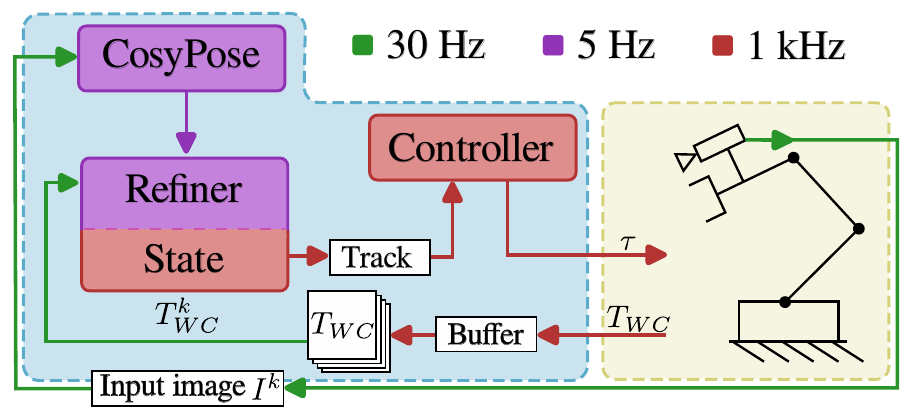}
    \caption{
    \textbf{The robot control architecture} used for the tracking experiment. First, an image $I^k$ is used with \textcolor{blue-violet}{CosyPose} to generate object pose estimates. These estimates are then fed into the proposed \textcolor{blue-violet}{Refiner} along with the camera pose $T_{WC}^k$ whose timestamp corresponds to the time stamp of the input image $I^k$ used in CosyPose. This synchronization is achieved by buffering the poses $T_{WC}$ and subsequently selecting the one with the closest timestamp. The Refiner produces an estimate of the \textcolor{bostonuniversityred}{State}, \ie, the probabilistic world model.
    Note that although the world model is updated at CosyPose frequency, the \textcolor{bostonuniversityred}{State} is computed at the robot control frequency using the motion model.  
    Finally, a track selected by the user is used as input for the robot \textcolor{bostonuniversityred}{Controller}, which computes the motor torques $\pmb \tau$ required to move the robot into the desired pose.
    The typical processing frequencies of individual modules are  \textcolor{blue-violet}{5 Hz} for CosyPose and Refiner,  \textcolor{ao(english)}{30~Hz} for the camera, and \textcolor{bostonuniversityred}{1~kHz} for the state extrapolation and robot controller.
    }
    \label{fig:experiment_diagram}
    \vspace{-5mm}    
\end{figure}

\begin{figure}[htb]
    \centering
    \includegraphics[width=\linewidth]{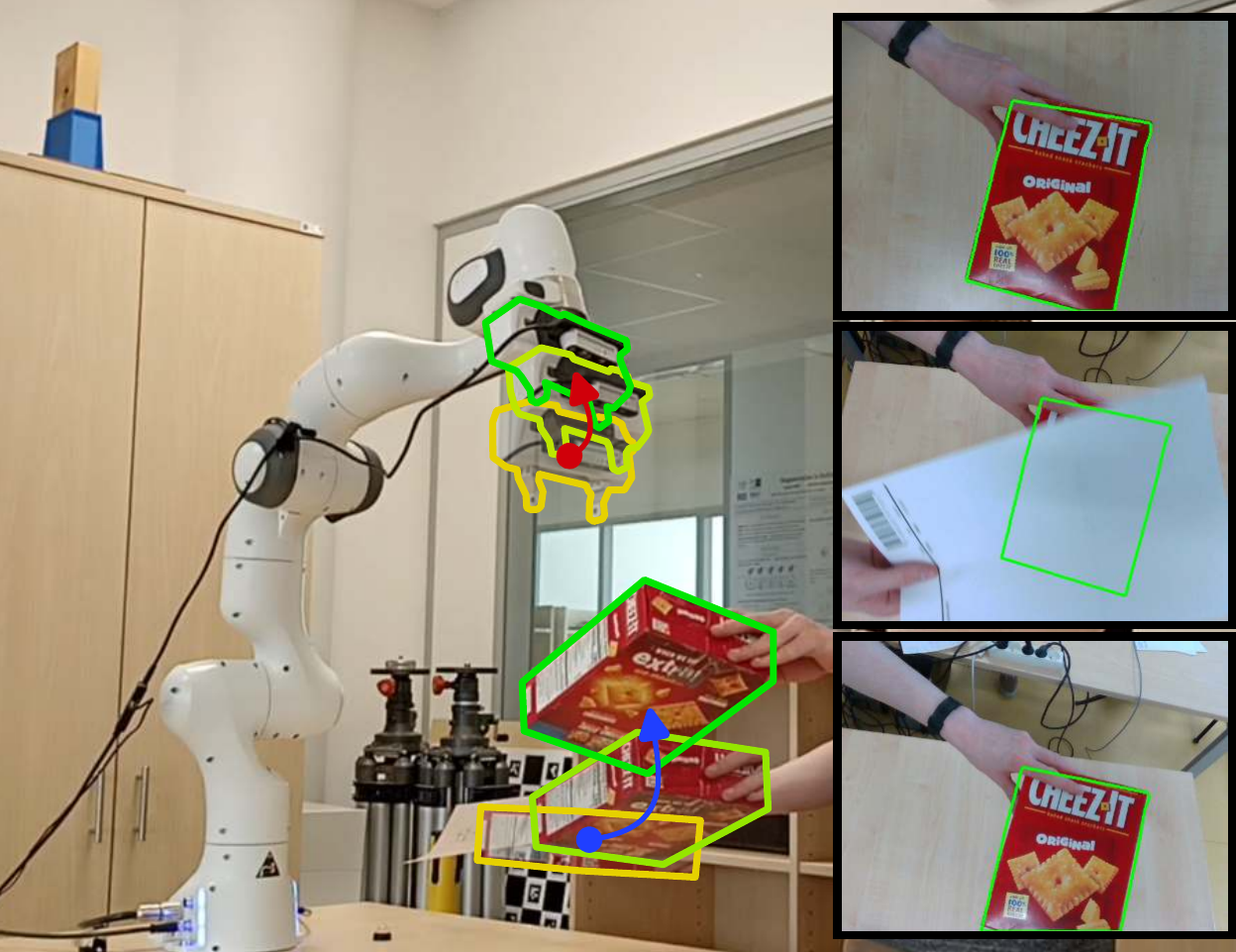}
    \caption{
    \textbf{Robot tracking experiment.}
    The illustration depicts a selected sequence of images recorded during an experiment where the robot attempts to maintains a constant relative end-effector transformation with respect to the Cheez-it box from the YCB~\cite{calli2015ycb} dataset. During the tracking process, the object is occluded by a sheet of paper, demonstrating the temporal consistency and stability of the refined pose estimates.
    }
    \label{fig:tracking_demo}
    \vspace{-0mm}    
\end{figure}

For the dynamic object tracking experiment, we implement the Cartesian impedance control~\cite{khatib1987unified} using the estimated target object pose as reference.
The controller architecture is visualized in Fig.~\ref{fig:experiment_diagram}.
With this control architecture and using the proposed filtering method, we achieve stable tracking in a challenging scenario in which the object is hidden behind an occluder, as shown in Fig.~\ref{fig:tracking_demo}. 
The analysis of the image stream for the tracking experiment is shown in Fig.~\ref{fig:tracking_analysis}.
The full experiment is shown in the supplementary video.

\begin{figure}[htb]
    \centering
    \includegraphics[width=\linewidth]{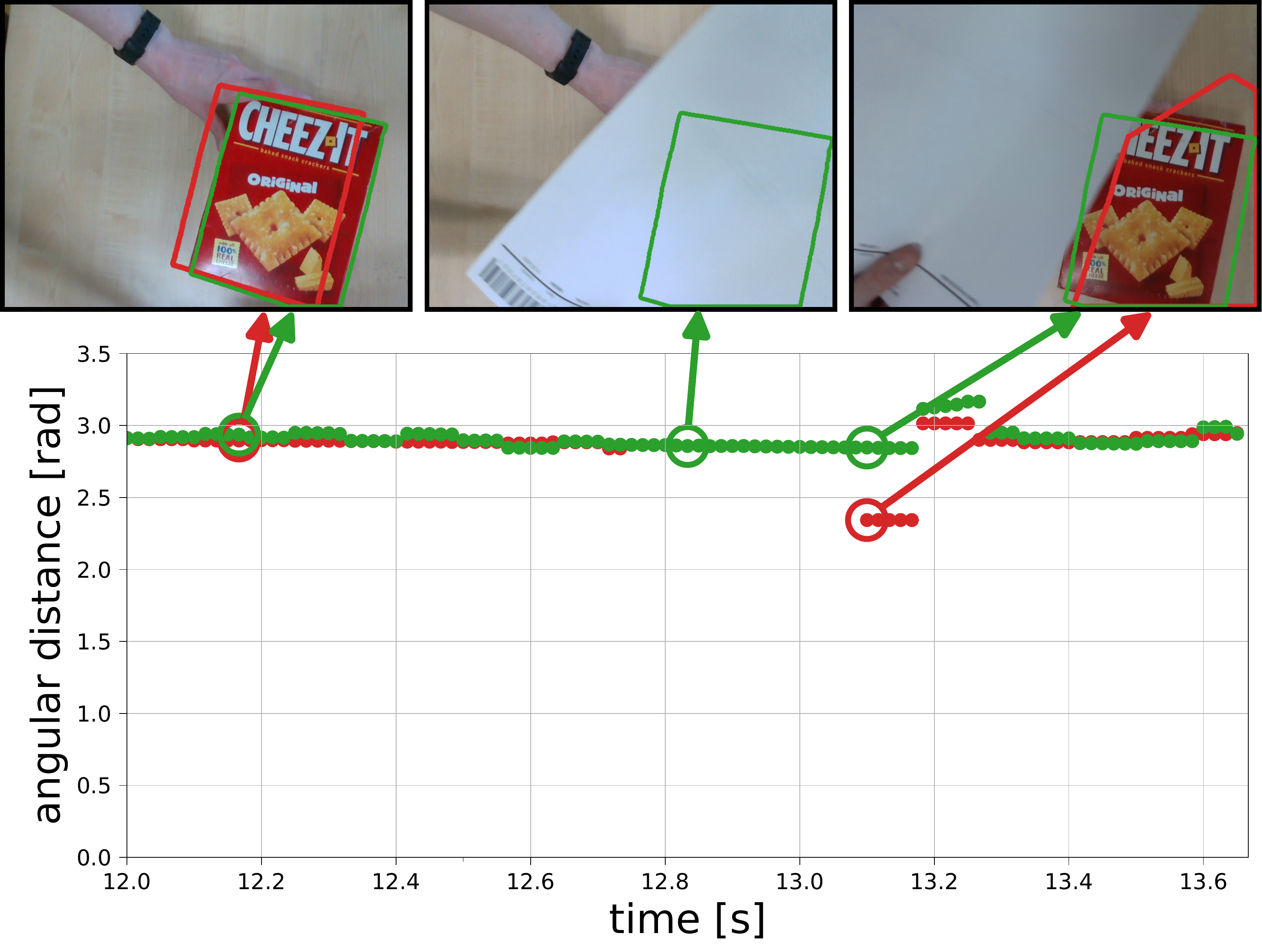}
    \caption{
    \textbf{Analysis of robot tracking experiment.}
    The evolution of the object angular distance for the robot tracking experiment.
    If the object is not occluded, CosyPose and our method predicts the object pose accurately (first frame).
    However, when object is completely occluded the per-frame evaluation cannot evaluate the pose of the object (second frame).
    Finally, if the object is partially visible, CosyPose predicts wrong orientation while the proposed estimator remains stable (third frame).
    }
    \label{fig:tracking_analysis}
    \vspace{-0mm}
\end{figure}

\section{CONCLUSION}
Accurate and temporally consistent object pose estimation is crucial for robot interaction with both static and dynamically moving objects.
This work demonstrates that it is beneficial to consider the full stream of images rather than the per-frame estimates to achieve robust temporally smooth predictions.
The proposed algorithm for probabilistic filtering has been validated both quantitatively on three benchmarks and qualitatively by tracking experiments involving a Panda robot, showing improved results while running in real time.

\noindent\textbf{Limitations.}
This work addressed object symmetries as separate tracks in the factor graph.
Although it works for discrete symmetries (\eg a box), continuous symmetries (\eg a cylinder) would create many low-confidence tracks that would be difficult to use for robot control. This may be addressed by assuming known symmetries and modifying the object pose factor, which is left for future work.

\bibliographystyle{IEEEtran}
\bibliography{references}


\end{document}